\documentclass{article}
\usepackage{iclr2020_conference,times}
\usepackage{latexsym}
\usepackage{tabularx}
\usepackage{graphicx}
\usepackage{float}
\usepackage{enumerate}
\usepackage{amssymb,amsmath}
\usepackage{color}
\usepackage{bm}
\usepackage{makecell}
\usepackage{multi row}
\usepackage{array}
\usepackage{resizegather}
\usepackage{refcount}
\usepackage{fixfoot}
\usepackage{hyperref}
\usepackage{url}
\usepackage{subcaption} 
\usepackage{bbm}
\usepackage{listings}

\newif\ifcomment
\commenttrue
\ifcomment
\newcommand{\kc}[1]{\textcolor{red}{KC: #1}}
\newcommand{\tl}[1]{\textcolor{blue}{TL: #1}}
\newcommand{\ql}[1]{\textcolor{cyan}{QL: #1}}
\newcommand{\cm}[1]{\textcolor{orange}{CM: #1}}
\definecolor{CMpurple}{rgb}{0.6,0.18,0.64}
\newcommand\cmm[1]{\marginpar{\tiny\raggedright\textcolor{CMpurple}{\textsf{\bfseries CM\@: #1}}}}
\else
\newcommand{\kc}[1]{}
\newcommand{\tl}[1]{}
\newcommand{\ql}[1]{}
\newcommand{\cm}[1]{}
\newcommand{\cmm}[1]{}
\fi

\newcommand\tstrut{\rule{0pt}{2.6ex}}         
\newcommand\bstrut{\rule[-1.0ex]{0pt}{0pt}}   
\newcommand{\thinline}{\Xhline{1.5\arrayrulewidth}}
\newcommand{\thickline}{\Xhline{2.5\arrayrulewidth}}
\newcommand{\tsep}	{\bstrut \\ \thinline}

\newcommand{\ttop}{\thickline}
\newcommand{\tbottom}{\bstrut \\ \thickline}

\newcommand{\xhdr}[1]{\vspace{0mm}\noindent{{\bf #1}}\hspace{1.3mm}}

\newcommand{\alns}[1] {
	\begin{align*} #1 \end{align*}
}
\newcommand{\pdata}{p_{\text{data}}}
\newcommand{\pmask}{p_{\text{mask}}}
\newcommand{\bmx}{\bm{x}^\text{masked}}
\newcommand{\bfx}{\bm{x}^\text{corrupt}}

\newcommand{\fx}{x^\text{corrupt}}
\newcommand{\bx}{\bm{x}}

\newcommand{\hbx}{\hat{\bm{x}}}
\newcommand{\E} {\mathop{\mathbb{E}}}

\newcommand{\D}{D}
\newcommand{\I}{\bm{m}}
\newcommand{\lossg}{\mathcal{L}_\text{MLM}(\bx, \theta_G)}
\newcommand{\lossd}{\mathcal{L}_\text{Disc}(\bx, \theta_D)}
\newcommand{\lossgs}{\mathcal{L}_\text{MLM}}
\newcommand{\lossds}{\mathcal{L}_\text{Disc}}
\newcommand{\replace}[3] {
    \textsc{replace}(#1, #2, #3)
}

\DeclareMathOperator*{\argmax}{arg\,max}

\title{ELECTRA: Pre-training Text Encoders\\ as Discriminators Rather Than Generators}

\author{Kevin Clark\\Stanford University\\{\tt kevclark@cs.stanford.edu}
\And Minh-Thang Luong\\Google Brain\\{\tt thangluong@google.com}
\And Quoc V. Le\\Google Brain\\{\tt qvl@google.com}
\And Christopher D. Manning\\Stanford University \& CIFAR Fellow\\{\tt manning@cs.stanford.edu}
}

\iclrfinalcopy 
\begin{document}

\maketitle

\begin{abstract}
Masked language modeling (MLM) pre-training methods such as BERT corrupt the input by replacing some tokens with \texttt{[MASK]} and then train a model to reconstruct the original tokens.
While they produce good results when transferred to downstream NLP tasks, they generally require large amounts of compute to be effective.
As an alternative, we propose a more sample-efficient pre-training task called replaced token detection.
Instead of masking the input, our approach corrupts it by replacing some tokens with plausible alternatives sampled from a small generator network.
Then, instead of training a model that predicts the original identities of the corrupted tokens, we train a discriminative model that predicts whether each token in the corrupted input was replaced by a generator sample or not.
Thorough experiments demonstrate this new pre-training task is more efficient than MLM because the task is defined over {\it all} input tokens rather than just the small subset that was masked out. 
As a result, the contextual representations learned by our approach substantially outperform the ones learned by BERT given the same model size, data, and compute.
The gains are particularly strong for small models; for example, we train a model on one GPU for 4 days that outperforms GPT (trained using 30x more compute) on the GLUE natural language understanding benchmark.
Our approach also works well at scale, where it performs comparably to RoBERTa and XLNet while using less than 1/4 of their compute and outperforms them when using the same amount of compute. 
\end{abstract}

\section{Introduction}

Current state-of-the-art representation learning methods for language can be viewed as learning denoising autoencoders \citep{vincent2008extracting}.
They select a small subset of the unlabeled input sequence (typically 15\%), mask the identities of those tokens (e.g., BERT; \citet{devlin2018bert}) or attention to those tokens (e.g., XLNet; \citet{yang2019xlnet}), and then train the network to recover the original input.
While more effective than conventional language-model pre-training due to learning bidirectional representations, these masked language modeling (MLM) approaches incur a substantial compute cost because the network only learns from 15\% of the tokens per example.

As an alternative, we propose \emph{replaced token detection,} a pre-training task in which the model learns to distinguish real input tokens from plausible but synthetically generated replacements. 
Instead of masking, our method corrupts the input by replacing some tokens with samples from a proposal distribution, which is typically the output of a small masked language model.
This corruption procedure solves a mismatch in BERT (although not in XLNet) where the network sees artificial $\texttt{[MASK]}$ tokens during pre-training but not when being fine-tuned on downstream tasks.
We then pre-train the network as a discriminator that predicts for every token whether it is an original or a replacement. 
In contrast, MLM trains the network as a generator that predicts the original identities of the corrupted tokens.
A key advantage of our discriminative task is that the model learns from \emph{all} input tokens instead of just the small masked-out subset, making it more computationally efficient.
Although our approach is reminiscent of training the discriminator of a GAN, our method is not adversarial in that the generator producing corrupted tokens is trained with maximum likelihood due to the difficulty of applying GANs to text \citep{Caccia2018LanguageGF}.

We call our approach ELECTRA\footnote{Code and pre-trained weights will be released at \url{https://github.com/google-research/electra}} for ``Efficiently Learning an Encoder that Classifies Token Replacements Accurately." 
As in prior work, we apply it to pre-train Transformer text encoders \citep{Vaswani2017AttentionIA} that can be fine-tuned on downstream tasks.
Through a series of ablations, we show that learning from all input positions causes ELECTRA to train much faster than BERT.
We also show ELECTRA achieves higher accuracy on downstream tasks when fully trained.  

\begin{figure}[tb]
\begin{center}
\includegraphics[width=1.0\textwidth]{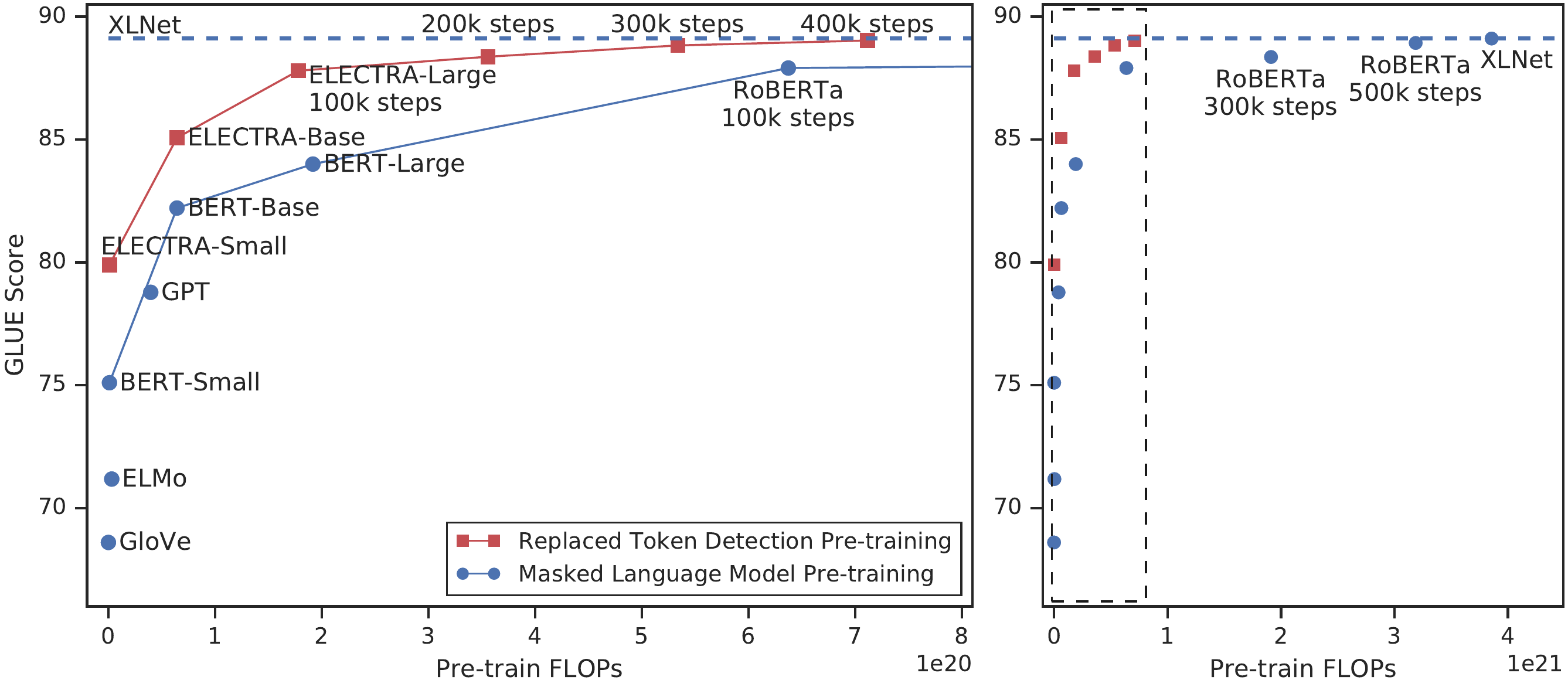}
\end{center}
\caption{Replaced token detection pre-training consistently outperforms masked language model pre-training given the same compute budget. The left figure is a zoomed-in view of the dashed box.}
\label{fig:main}
\end{figure}

Most current pre-training methods require large amounts of compute to be effective, raising concerns about their cost and accessibility.
Since pre-training with more compute almost always results in better downstream accuracies, we argue an important consideration for pre-training methods should be compute efficiency as well as absolute downstream performance.
From this viewpoint, we train ELECTRA models of various sizes and evaluate their downstream performance vs.\ their compute requirement.
In particular, we run experiments on the GLUE natural language understanding benchmark \citep{wang2018glue} and SQuAD question answering benchmark \citep{Rajpurkar2016SQuAD10}.
ELECTRA substantially outperforms MLM-based methods such as BERT and XLNet given the same model size, data, and compute (see Figure~\ref{fig:main}).
For example, we build an ELECTRA-Small model that can be trained on 1 GPU in 4 days.\footnote{It has 1/20th the parameters and requires 1/135th the pre-training compute of BERT-Large.}
ELECTRA-Small outperforms a comparably small BERT model by 5 points on GLUE, and even outperforms the much larger GPT model \citep{radford2018improving}.
Our approach also works well at large scale, where we train an ELECTRA-Large model that performs comparably to RoBERTa \citep{liu2019roberta} and XLNet \citep{yang2019xlnet}, despite having fewer parameters and using 1/4 of the compute for training.
Training ELECTRA-Large further results in an even stronger model that outperforms ALBERT \citep{lan2019albert} on GLUE and sets a new state-of-the-art for SQuAD 2.0.  
Taken together, our results indicate that the discriminative task of distinguishing real data from challenging negative samples is more compute-efficient and parameter-efficient than existing generative approaches for
language representation learning.

\section{Method}
\label{sec:method}

\begin{figure}[tb]
\begin{center}
\includegraphics[width=0.95\textwidth]{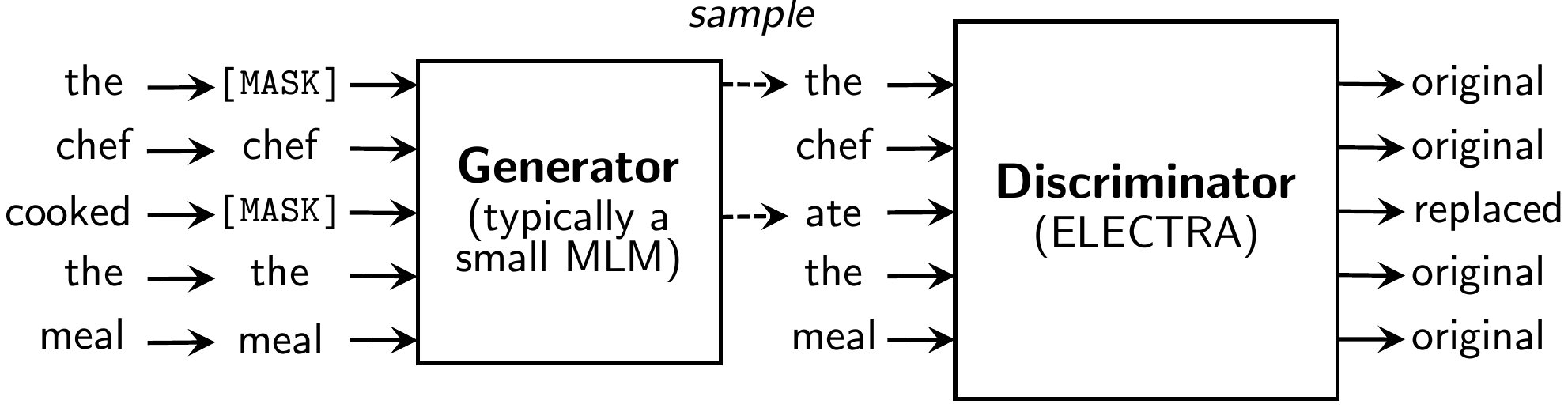}
\end{center}
\caption{An overview of replaced token detection. The generator can be any model that produces an output distribution over tokens, but we usually use a small masked language model that is trained jointly with the discriminator.
Although the models are structured like in a GAN, we train the generator with maximum likelihood rather than adversarially due to the difficulty of applying GANs to text.
After pre-training, we throw out the generator and only fine-tune the discriminator (the ELECTRA model) on downstream tasks.
}
\label{fig:overview}
\end{figure}

We first describe the  replaced token detection pre-training task; see Figure~\ref{fig:overview} for an overview.
We suggest and evaluate several modeling improvements for this method in Section~\ref{sec:extensions}.

Our approach trains two neural networks, a generator $G$ and a discriminator $D$.
Each one primarily consists of an encoder (e.g., a Transformer network) that maps a sequence on input tokens $\bx = [x_1, ..., x_n]$ into a sequence of contextualized vector representations
$h(\bx) = [h_1, ..., h_n]$.
For a given position $t$, (in our case only positions where $x_t = \texttt{[MASK]}$), 
the generator outputs a probability for generating a particular token $x_t$ with a softmax layer: 
\alns{
 p_G(x_t | \bx) = \text{exp}\left(e(x_t)^T h_G(\bx)_t\right) / \sum_{x'} \exp\left(e(x')^T h_G(\bx)_t\right)
} 
where $e$ denotes token embeddings. 
For a given position $t$, the discriminator predicts whether the token $x_t$ is ``real," i.e., that it comes from the data rather than the generator distribution, with a sigmoid output layer: 
\alns{ 
    D(\bx, t) = \text{sigmoid}(w^T h_D(\bx)_t)
}
The generator is trained to perform masked language modeling (MLM). Given an input $\bx = [x_1, x_2, ..., x_n]$, MLM first select a random set of positions (integers between 1 and $n$) to mask out $\I = [m_1, ..., m_k]$.\footnote{Typically $k = \lceil0.15n\rceil$, i.e., 15\% of the tokens are masked out.}
The tokens in the selected positions are replaced with a $\texttt{[MASK]}$ token: we denote this as $\bmx = \replace{\bx}{\I}{\texttt{[MASK]}}$.
The generator then learns to predict the original identities of the masked-out tokens. 
The discriminator is trained to distinguish tokens in the data from tokens that have been replaced by generator samples.
More specifically, we create a corrupted example $\bfx$ by replacing the masked-out tokens with generator samples and train the discriminator to predict which tokens in $\bfx$ match the original input $\bx$. Formally, model inputs are constructed according to
\alns{
&m_i \sim \text{unif}\{1, n\} \text{ for } i=1 \text{ to } k \hspace{10mm} 
\bmx = \replace{\bx}{\I}{\texttt{[MASK]}} \\
&\hat{x}_i \sim p_G(x_i | \bmx) \text{ for } i \in \I \hspace{10mm}
\bfx = \replace{\bx}{\I}{\hbx} 
}
and the loss functions are

\vspace{-1mm}
\resizebox{\linewidth}{!}{
  \begin{minipage}{\linewidth}
\alns{
&\lossg = \E \left( \sum_{i \in \I} -\log p_G(x_i | \bmx) \right) \\
&\lossd = \E \left(\sum_{t=1}^n -\mathbbm{1}(\fx_t = x_t)\log \D(\bfx, t) - \mathbbm{1}(\fx_t \neq x_t)\log(1 - \D(\bfx, t)) \right)
}
\end{minipage}
}

Although similar to the training objective of a GAN, there are several key differences.
First, if the generator happens to generate the correct token, that token is considered ``real" instead of ``fake"; we found this formulation to moderately improve results on downstream tasks.
More importantly, the generator is trained with maximum likelihood rather than being trained adversarially to fool the discriminator. 
Adversarially training the generator is challenging because it is impossible to back-propagate through sampling from the generator. 
Although we experimented circumventing this issue by using reinforcement learning to train the generator (see Appendix~\ref{app:adv}), this performed worse than maximum-likelihood training.
Lastly, we do not supply the generator with a noise vector as input, as is typical with a GAN. 

We minimize the combined loss
\alns{
    \min_{\theta_G, \theta_D} \sum_{\bx \in \mathcal{X}} \lossg + \lambda \lossd
}
over a large corpus $\mathcal{X}$ of raw text. We approximate the expectations in the losses with a single sample. 
We don't back-propagate the discriminator loss through the generator (indeed, we can't because of the sampling step). 
After pre-training, we throw out the generator and fine-tune the discriminator on downstream tasks.

\section{Experiments}

\subsection{Experimental Setup}

We evaluate on the General Language Understanding Evaluation (GLUE) benchmark \citep{wang2018glue} and Stanford Question Answering (SQuAD) dataset \citep{Rajpurkar2016SQuAD10}. GLUE contains a variety of tasks covering textual entailment (RTE and MNLI) question-answer entailment (QNLI), paraphrase (MRPC), question paraphrase (QQP), textual similarity (STS), sentiment (SST), and linguistic acceptability (CoLA). See Appendix ~\ref{app:glue} for more details on the GLUE tasks.
Our evaluation metrics are Spearman correlation for STS, Matthews correlation for CoLA, and accuracy for the other GLUE tasks; we generally report the average score over all tasks. 
For SQuAD, we evaluate on versions 1.1, in which models select the span of text answering a question, and 2.0, in which some questions are unanswerable by the passage. 
We use the standard evaluation metrics of Exact-Match (EM) and F1 scores.  
For most experiments we pre-train on the same data as BERT, which consists of 3.3 Billion tokens from Wikipedia and BooksCorpus \citep{Zhu2015AligningBA}. 
However, for our Large model we pre-trained on the data used for XLNet \citep{yang2019xlnet}, which extends the BERT dataset to 33B tokens by including data from ClueWeb \citep{callan2009clueweb09}, CommonCrawl, and Gigaword \citep{ParkerGiga}. 
All of the pre-training and evaluation is on English data, although we think it would be interesting to apply our methods to multilingual data in the future. 

Our model architecture and most hyperparameters are the same as BERT's. 
For fine-tuning on GLUE, we add simple linear classifiers on top of ELECTRA.
For SQuAD, we add the question-answering module from XLNet on top of ELECTRA, which is slightly more sophisticated than BERT's in that it jointly rather than independently predicts the start and end positions and has a ``answerability" classifier added for SQuAD 2.0.
Some of our evaluation datasets are small, which means accuracies of fine-tuned models can vary substantially depending on the random seed. 
We therefore report the median of 10 fine-tuning runs from the same pre-trained checkpoint for each result.
Unless stated otherwise, results are on the dev set.
See the appendix for further training details and hyperparameter values.

\subsection{Model Extensions} 
\label{sec:extensions}
We improve our method by proposing and evaluating several extensions to the model. Unless stated otherwise, these experiments use the same model size and training data as BERT-Base.

\xhdr{Weight Sharing}
We propose improving the efficiency of the pre-training by sharing weights between the generator and discriminator. 
If the generator and discriminator are the same size, all of the transformer weights can be tied.
However, we found it to be more efficient to have a small generator, in which case we only share the embeddings (both the token and positional embeddings) of the generator and discriminator.
In this case we use embeddings the size of the discriminator's hidden states.\footnote{We add linear layers to the generator to project the embeddings into generator-hidden-sized representations.}
The ``input" and ``output" token embeddings of the generator are always tied as in BERT. 

We compare the weight tying strategies when the generator is the same size as the discriminator. We train these models for 500k steps. GLUE scores are 83.6 for no weight tying, 84.3 for tying token embeddings, and 84.4 for tying all weights.
We hypothesize that ELECTRA benefits from tied token embeddings because masked language modeling is particularly effective at learning these representations: while the discriminator only updates tokens that are present in the input or are sampled by the generator, the generator's softmax over the vocabulary densely updates all token embeddings. 
On the other hand, tying all encoder weights caused little improvement while incurring the significant disadvantage of requiring the generator and discriminator to be the same size. 
Based on these findings, we use tied embeddings for further experiments in this paper. 

\begin{figure}[tb]
\begin{center}
\begin{minipage}[c]{0.55\textwidth}
\includegraphics[width=\textwidth]{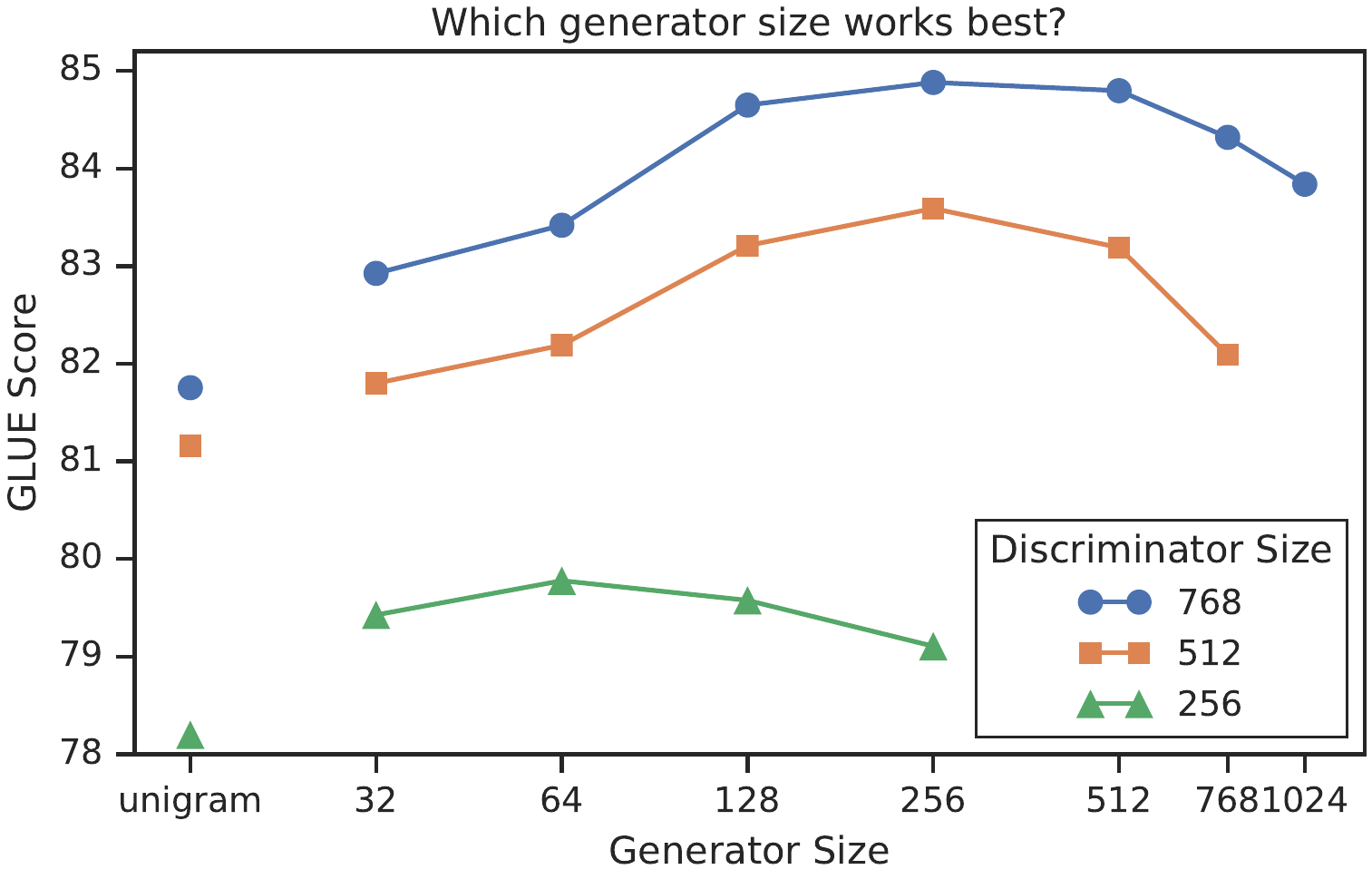}
\end{minipage}
\hspace{4mm}
\begin{minipage}[c]{0.4\textwidth}
\includegraphics[width=\textwidth]{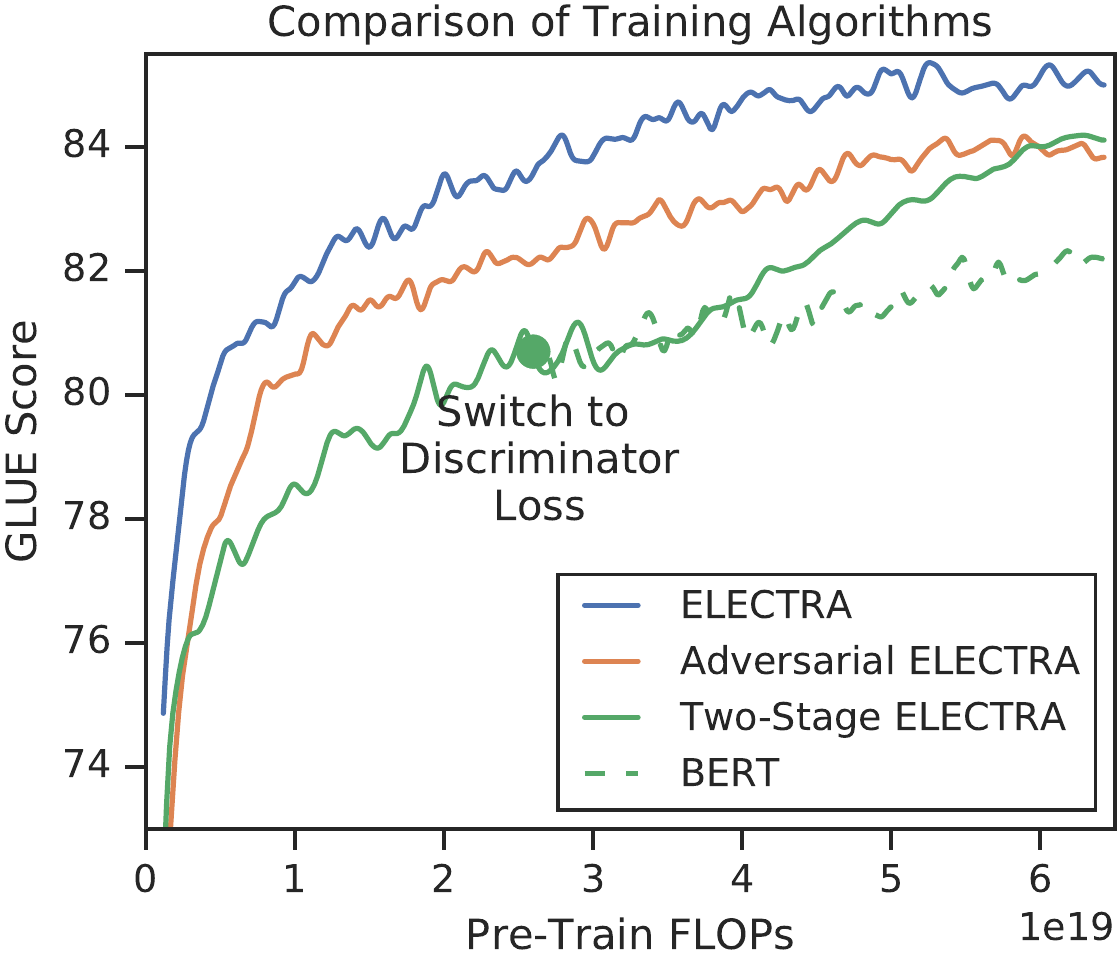}
\end{minipage}
\end{center}
\caption{\underline{Left}: GLUE scores for different generator/discriminator sizes (number of hidden units). Interestingly, having a generator smaller than the discriminator improves results.
\underline{\smash{Right}}: Comparison of different training algorithms. As our focus is on efficiency, the x-axis shows FLOPs rather than train steps (e.g., ELECTRA is trained for fewer steps than BERT because it includes the generator).}
\label{fig:extensions}
\end{figure}

\xhdr{Smaller Generators}
If the generator and discriminator are the same size, training ELECTRA would take around twice as much compute per step as training only with masked language modeling.
We suggest using a smaller generator to reduce this factor.
Specifically, we make models smaller by decreasing the layer sizes while keeping the other hyperparameters constant.
We also explore using an extremely simple ``unigram" generator that samples fake tokens according their frequency in the train corpus. 
GLUE scores for differently-sized generators and discriminators are shown in the left of Figure~\ref{fig:extensions}.
All models are trained for 500k steps, which puts the smaller generators at a disadvantage in terms of compute because they require less compute per training step.
Nevertheless, we find that models work best with generators 1/4-1/2 the size of the discriminator. 
We speculate that having too strong of a generator may pose a too-challenging task for the discriminator, preventing it from learning as effectively.
In particular, the discriminator may have to use many of its parameters modeling the generator rather than the actual data distribution.
Further experiments in this paper use the best generator size found for the given discriminator size.

\xhdr{Training Algorithms}
Lastly, we explore other training algorithms for ELECTRA, although these did not end up improving results.
The proposed training objective jointly trains the generator and discriminator. We experiment with instead using the following two-stage training procedure:
\begin{enumerate}
    \vspace{-1mm}
    \item Train only the generator with $\lossgs$ for $n$ steps.
    \vspace{-0.3mm}
    \item Initialize the weights of the discriminator with the weights of the generator. Then train the discriminator with $\lossds$ for $n$ steps, keeping the generator's weights frozen.
    \vspace{-1mm}
\end{enumerate}
Note that the weight initialization in this procedure requires having the same size for the generator and discriminator.
We found that without the weight initialization the discriminator would sometimes fail to learn at all beyond the majority class, perhaps because the generator started so far ahead of the discriminator. 
Joint training on the other hand naturally provides a curriculum for the discriminator where the generator starts off weak but gets better throughout training.
We also explored training the generator adversarially as in a GAN, using reinforcement learning to accommodate the discrete operations of sampling from the generator. See Appendix~\ref{app:adv} for details. 

Results are shown in the right of Figure~\ref{fig:extensions}.
During two-stage training, downstream task performance notably improves after the switch from the generative to the discriminative objective, but does not end up outscoring joint training. 
Although still outperforming BERT, we found adversarial training to underperform maximum-likelihood training.
Further analysis suggests the gap is caused by two problems with adversarial training.
First, the adversarial generator is simply worse at masked language modeling; it achieves 58\% accuracy at masked language modeling compared to 65\% accuracy for an MLE-trained one.
We believe the worse accuracy is mainly due to the poor sample efficiency of reinforcement learning when working in the large action space of generating text.
Secondly, the adversarially trained generator produces a low-entropy output distribution where most of the probability mass is on a single token, which means there is not much diversity in the generator samples. 
Both of these problems have been observed in GANs for text in prior work \citep{Caccia2018LanguageGF}.

\subsection{Small Models}

As a goal of this work is to improve the efficiency of pre-training, we develop a small model that can be quickly trained on a single GPU. 
Starting with the BERT-Base hyperparameters, we shortened the sequence length (from 512 to 128), reduced the batch size (from 256 to 128), reduced the model's hidden dimension size (from 768 to 256), and used smaller token embeddings (from 768 to 128).
To provide a fair comparison, we also train a BERT-Small model using the same hyperparameters.
We train BERT-Small for 1.5M steps, so it uses the same training FLOPs as ELECTRA-Small, which was trained for 1M steps.\footnote{ELECTRA requires more FLOPs per step because it consists of the generator as well as the discriminator.}
In addition to BERT, we compare against two less resource-intensive pre-training methods based on language modeling: ELMo \citep{peters2018deep} and GPT \citep{radford2018improving}.\footnote{GPT is similar in size to BERT-Base, but is trained for fewer steps.}
We also show results for a base-sized ELECTRA model comparable to BERT-Base. 

Results are shown in Table~\ref{tab:small}. See Appendix~\ref{app:test} for additional results, including stronger small-sized and base-sized models trained with more compute. ELECTRA-Small performs remarkably well given its size, achieving a higher GLUE score than other methods using substantially more compute and parameters.
For example, it scores 5 points higher than a comparable BERT-Small model and even outperforms the much larger GPT model. 
ELECTRA-Small is trained mostly to convergence, with models trained for even less time (as little as 6 hours) still achieving reasonable performance.
While small models distilled from larger pre-trained transformers can also achieve good GLUE scores \citep{sunmobilebert,jiao2019tinybert}, these models require first expending substantial compute to pre-train the larger teacher model.
The results also demonstrate the strength of ELECTRA at a moderate size; our base-sized ELECTRA model substantially outperforms BERT-Base and even outperforms BERT-Large (which gets 84.0 GLUE score). 
We hope ELECTRA's ability to achieve strong results with relatively little compute will broaden the accessibility of developing and applying pre-trained models in NLP.

\addtolength{\tabcolsep}{-1.2pt}
\begin{table*}[t!]
\small
\begin{center}
\begin{tabularx}{\linewidth}{X l l l l l}
\ttop
\textbf{Model} & \textbf{Train / Infer FLOPs} & \textbf{Speedup} & \textbf{Params} & \textbf{Train Time + Hardware} & \textbf{GLUE} \tstrut \tsep
ELMo & 3.3e18 / 2.6e10  & 19x / 1.2x & 96M & 14d on 3 GTX 1080 GPUs & 71.2 \tstrut \\
GPT & 4.0e19 / 3.0e10 & 1.6x / 0.97x & 117M & 25d on 8 P6000 GPUs & 78.8 \\
BERT-Small & 1.4e18 / 3.7e9 & 45x / 8x  & 14M & 4d on 1 V100 GPU  & 75.1 \\
BERT-Base & 6.4e19 / 2.9e10 & 1x / 1x & 110M & 4d on 16 TPUv3s &  82.2 \tsep
ELECTRA-Small & 1.4e18 / 3.7e9 & 45x / 8x & 14M & 4d on 1 V100 GPU & 79.9 \tstrut \\
\hspace{3mm}50\% trained & 7.1e17 / 3.7e9 & 90x / 8x & 14M & 2d on 1 V100 GPU & 79.0 \\
\hspace{3mm}25\% trained & 3.6e17 / 3.7e9 & 181x / 8x & 14M & 1d on 1 V100 GPU & 77.7 \\
\hspace{3mm}12.5\% trained & 1.8e17 / 3.7e9 & 361x / 8x & 14M & 12h on 1 V100 GPU & 76.0 \\
\hspace{3mm}6.25\% trained & 8.9e16 / 3.7e9 & 722x / 8x & 14M & 6h on 1 V100 GPU & 74.1 \\
ELECTRA-Base & 6.4e19 / 2.9e10 & 1x / 1x & 110M & 4d on 16 TPUv3s & 85.1 \tbottom
\end{tabularx} 
\end{center}
\vspace{-1mm}
\caption{Comparison of small models on the GLUE dev set.
BERT-Small/Base are our implementation and use the same hyperparameters as ELECTRA-Small/Base. 
Infer FLOPs assumes single length-128 input. 
Training times should be taken with a grain of salt as they are for different hardware and with sometimes un-optimized code.
ELECTRA performs well even when trained on a single GPU, scoring 5 GLUE points higher than a comparable BERT model and even outscoring the much larger GPT model.
}
\label{tab:small}
\end{table*}
\addtolength{\tabcolsep}{1.2pt}
\vspace{-0.0mm}

\subsection{Large Models}
We train big ELECTRA models to measure the effectiveness of the replaced token detection pre-training task at the large scale of current state-of-the-art pre-trained Transformers.
Our ELECTRA-Large models are the same size as BERT-Large but are trained for much longer.
In particular, we train a model for 400k steps (ELECTRA-400K; roughly 1/4 the pre-training compute of RoBERTa) and one for 1.75M steps (ELECTRA-1.75M; similar compute to RoBERTa).
We use a batch size 2048 and the XLNet pre-training data.
We note that although the XLNet data is similar to the data used to train RoBERTa, the comparison is not entirely direct. 
As a baseline, we trained our own BERT-Large model using the same hyperparameters and training time as ELECTRA-400K.

Results on the GLUE dev set are shown in Table~\ref{tab:large}. 
ELECTRA-400K performs comparably to RoBERTa and XLNet. 
However, it took less than 1/4 of the compute to train ELECTRA-400K as it did to train RoBERTa and XLNet, demonstrating that ELECTRA's sample-efficiency gains hold at large scale.
Training ELECTRA for longer (ELECTRA-1.75M) results in a model that outscores them on most GLUE tasks while still requiring less pre-training compute.
Surprisingly, our baseline BERT model scores notably worse than RoBERTa-100K, suggesting our models may benefit from more hyperparameter tuning or using the RoBERTa training data.
ELECTRA's gains hold on the GLUE test set (see Table~\ref{tab:test}), although these comparisons are less apples-to-apples due to the additional tricks employed by the models (see Appendix~\ref{app:fine}).  \\

\addtolength{\tabcolsep}{-3.5pt}
\begin{table*}[t!]
\small
\begin{center}
\begin{tabularx}{\linewidth}{X l l l l l l l l l l l}
\ttop 
\textbf{Model} & \textbf{Train FLOPs} & \textbf{Params} & \textbf{CoLA} & \textbf{SST} & \textbf{MRPC} & \textbf{STS} & \textbf{QQP} & \textbf{MNLI} & \textbf{QNLI} & \textbf{RTE} & \textbf{Avg.} \tstrut \tsep 
BERT      &  1.9e20 (0.27x)  & 335M  & 60.6 & 93.2 & 88.0 & 90.0 & 91.3 & 86.6 & 92.3 & 70.4 & 84.0 \tstrut \\
RoBERTa-100K      &  6.4e20 (0.90x) & 356M    & 66.1 & 95.6 & \textbf{91.4} & 92.2 & 92.0 & 89.3 & 94.0 & 82.7& 87.9  \\
RoBERTa-500K      &  3.2e21 (4.5x) & 356M   & 68.0 & 96.4 & 90.9 & 92.1 & 92.2 & 90.2 & 94.7 & 86.6 & 88.9 \\
XLNet      & 3.9e21 (5.4x) &  360M  & 69.0 & \textbf{97.0} & 90.8 & 92.2 & 92.3 & 90.8 & 94.9 & 85.9 & 89.1 \tsep
BERT (ours) &  7.1e20 (1x) & 335M  & 67.0 & 95.9 & 89.1 & 91.2 & 91.5 & 89.6 & 93.5 & 79.5 & 87.2  \tstrut \\
ELECTRA-400K      &  7.1e20 (1x) & 335M    & \textbf{69.3} & 96.0 & 90.6 & 92.1 & \textbf{92.4} & 90.5 & 94.5 & 86.8 & 89.0 \\ 
ELECTRA-1.75M      &  3.1e21 (4.4x) & 335M    & 69.1 & 96.9 & 90.8 & \textbf{92.6} & \textbf{92.4} & \textbf{90.9} & \textbf{95.0} & \textbf{88.0} & \textbf{89.5} \tbottom

\end{tabularx} 
\end{center}
\vspace{-1mm}
\caption{Comparison of large models on the GLUE dev set. ELECTRA and RoBERTa are shown for different numbers of pre-training steps, indicated by the numbers after the dashes. ELECTRA performs comparably to XLNet and RoBERTa when using less than 1/4 of their pre-training compute and outperforms them when given a similar amount of pre-training compute. BERT dev results are from \citet{Clark2019BAMBM}.}
\label{tab:large}
\end{table*}
\addtolength{\tabcolsep}{3.5pt}
\vspace{-1mm}

\addtolength{\tabcolsep}{-3.5pt}
\begin{table*}[t!]
\small
\begin{center}
\begin{tabularx}{\linewidth}{X l l l l l l l l l l l l}
\ttop 
\textbf{Model} & \textbf{Train FLOPs} & \textbf{CoLA} & \textbf{SST} & \textbf{MRPC} & \textbf{STS} & \textbf{QQP} & \textbf{MNLI} & \textbf{QNLI} & \textbf{RTE} & \textbf{WNLI} & \textbf{Avg.*} & \textbf{Score} \tstrut \tsep 
BERT      &  1.9e20 (0.06x)   & 60.5 & 94.9 & 85.4 & 86.5 & 89.3 & 86.7 & 92.7 & 70.1 & 65.1 & 79.8 & 80.5 \tstrut \\
RoBERTa      &  3.2e21 (1.02x)   & 67.8 & 96.7 & 89.8 & 91.9 & 90.2 & 90.8 & 95.4 & 88.2 & 89.0 & 88.1 & 88.1 \\
ALBERT      &  3.1e22 (10x)  & 69.1 & \textbf{97.1} & \textbf{91.2} & 92.0 & 90.5 & \textbf{91.3} & -- & 89.2 & 91.8 & 89.0 & -- \\
XLNet      & 3.9e21 (1.26x)    & 70.2 & \textbf{97.1} & 90.5 & \textbf{92.6} & 90.4 & 90.9 & -- & 88.5 & \textbf{92.5} & 89.1 & -- \tsep
ELECTRA &  3.1e21 (1x) & \textbf{71.7} & \textbf{97.1} & 90.7 & 92.5 & \textbf{90.8} & \textbf{91.3} & \textbf{95.8} & \textbf{89.8} & \textbf{92.5} & \textbf{89.5} & \textbf{89.4} \tstrut \tbottom
\end{tabularx} 
\end{center}
\vspace{-1mm}
\caption{GLUE test-set results for large models. 
Models in this table incorporate additional tricks such as ensembling to improve scores (see Appendix~\ref{app:fine} for details).
Some models do not have QNLI scores because they treat QNLI as a ranking task, which has recently been disallowed by the GLUE benchmark. To compare against these models, we report the average score excluding QNLI (Avg.*) in addition to the GLUE leaderboard score (Score). 
``ELECTRA" and ``RoBERTa" refer to the fully-trained ELECTRA-1.75M and RoBERTa-500K models.
}
\label{tab:test}
\end{table*}
\addtolength{\tabcolsep}{3.5pt}
\vspace{-1mm}

\addtolength{\tabcolsep}{-4.5pt}
\begin{table*}[t!]
\small
\begin{center}
\begin{tabularx}{1.0\linewidth}{X@{\hskip 4mm} l@{\hskip 4mm} l@{\hskip 4mm} c c c c c c}
\ttop
\multirow{2}{*}{\textbf{Model}} & \multirow{2}{*}{\textbf{Train FLOPs}} & \multirow{2}{*}{\textbf{Params}} & \multicolumn{2}{c}{\textbf{SQuAD 1.1 dev\phantom{....}}} & \multicolumn{2}{c}{\textbf{SQuAD 2.0 dev\phantom{....}}} & \multicolumn{2}{c}{\textbf{SQuAD 2.0 test}} \tstrut \\
 & & & EM & F1 & EM & F1 & EM & F1 \tsep
 BERT-Base & 6.4e19 (0.09x) & 110M &  80.8 & 88.5 & -- & -- & -- & -- \tstrut \\
 BERT & 1.9e20 (0.27x) & 335M &  84.1 & 90.9 & 79.0 & 81.8 & 80.0 & 83.0\\
 SpanBERT & 7.1e20 (1x) & 335M & 88.8 & 94.6 & 85.7 & 88.7 & 85.7 & 88.7 \\
 XLNet-Base & 6.6e19 (0.09x) & 117M & 81.3 & -- & 78.5 & -- & -- & -- \\
 XLNet & 3.9e21 (5.4x) & 360M & \textbf{89.7} & \textbf{95.1} & 87.9 & \textbf{90.6} & 87.9 & 90.7 \\
 RoBERTa-100K & 6.4e20 (0.90x) & 356M & -- & 94.0 & -- & 87.7 & -- & -- \\
 RoBERTa-500K & 3.2e21 (4.5x) & 356M & 88.9 & 94.6 & 86.5 & 89.4 & 86.8 & 89.8 \\
 ALBERT & 3.1e22 (44x) & 235M & 89.3 & 94.8 & 87.4 & 90.2& 88.1 & 90.9 \tsep
 BERT (ours) &  7.1e20 (1x) & 335M & 88.0 & 93.7 & 84.7 & 87.5& -- & -- \tstrut \\
 ELECTRA-Base & 6.4e19 (0.09x) & 110M & 84.5 & 90.8 & 80.5 & 83.3 & -- & --\\
 ELECTRA-400K & 7.1e20 (1x) & 335M & 88.7 & 94.2 & 86.9 & 89.6 & -- & --\\
 ELECTRA-1.75M & 3.1e21 (4.4x) & 335M & \textbf{89.7} & 94.9 & \textbf{88.0} & \textbf{90.6} & \textbf{88.7} & \textbf{91.4}
 \tbottom
\end{tabularx} 
\end{center}
\caption{Results on the SQuAD for non-ensemble models.}
\label{tab:squad}
\end{table*}
\addtolength{\tabcolsep}{4.5pt}

Results on SQuAD are shown in Table~\ref{tab:squad}. Consistent, with the GLUE results, ELECTRA scores better than masked-language-modeling-based methods given the same compute resources. 
For example, ELECTRA-400K outperforms RoBERTa-100k and our BERT baseline, which use similar amounts of pre-training compute. 
ELECTRA-400K also performs comparably to RoBERTa-500K despite using less than 1/4th of the compute. 
Unsurprisingly, training ELECTRA longer improves results further: ELECTRA-1.75M scores higher than previous models on the SQuAD 2.0 benchmark. 
ELECTRA-Base also yields strong results, scoring substantially better than BERT-Base and XLNet-Base, and even surpassing  BERT-Large according to most metrics. 
ELECTRA generally performs better at SQuAD 2.0 than 1.1. 
Perhaps replaced token detection, in which the model distinguishes real tokens from plausible fakes, is particularly transferable to the answerability classification of SQuAD 2.0, in which the model must distinguish answerable questions from fake unanswerable questions.

\subsection{Efficiency Analysis}
We have suggested that posing the training objective over a small subset of tokens makes masked language modeling inefficient.
However, it isn't entirely obvious that this is the case.
After all, the model still receives a large number of input tokens even though it predicts only a small number of masked tokens.
To better understand where the gains from ELECTRA are coming from, we compare a series of other pre-training objectives that are designed to be a set of ``stepping stones" between BERT and ELECTRA.
\begin{itemize}
    \item \textbf{ELECTRA 15\%}: This model is identical to ELECTRA except the discriminator loss only comes from the 15\% of the tokens that were masked out of the input. In other words, the sum in the discriminator loss $\lossds$ is over $i \in \I$ instead of from 1 to $n$.\footnote{We also trained a discriminator that learns from a random 15\% of the input tokens distinct from the subset that was originally masked out; this model performed slightly worse.}
    \item \textbf{Replace MLM}: This objective is the same as masked language modeling except instead of replacing masked-out tokens with $\texttt{[MASK]}$, they are replaced with tokens from a generator model. This objective tests to what extent ELECTRA's gains come from solving the discrepancy of exposing the model to $\texttt{[MASK]}$ tokens during pre-training but not fine-tuning.
    \item \textbf{All-Tokens MLM}: Like in Replace MLM, masked tokens are replaced with generator samples. Furthermore, the model predicts the identity of all tokens in the input, not just ones that were masked out. 
    We found it improved results to train this model with an explicit copy mechanism that outputs a copy probability $D$ for each token using a sigmoid layer. The model's output distribution puts $D$ weight on the input token plus $1 - D$ times the output of the MLM softmax. This model is essentially a combination of BERT and ELECTRA.
    Note that without generator replacements, the model would trivially learn to make predictions from the vocabulary for $\texttt{[MASK]}$ tokens and copy the input for other ones. 
\end{itemize}
Results are shown in Table~\ref{tab:sample}. 
First, we find that ELECTRA is greatly benefiting from having a loss defined over all input tokens rather than just a subset: ELECTRA 15\% performs much worse than ELECTRA. 
Secondly, we find that BERT performance is being slightly harmed from the pre-train fine-tune mismatch from $\texttt{[MASK]}$ tokens, as Replace MLM slightly outperforms BERT. 
We note that BERT (including our implementation) already includes a trick to help with the pre-train/fine-tune discrepancy: masked tokens are replaced with a random token 10\% of the time and are kept the same 10\% of the time. However, our results suggest these simple heuristics are insufficient to fully solve the issue. 
Lastly, we find that All-Tokens MLM, the generative model that makes predictions over all tokens instead of a subset, closes most of the gap between BERT and ELECTRA.
In total, these results suggest a large amount of ELECTRA's improvement can be attributed to learning from all tokens and a smaller amount can be attributed to alleviating the pre-train fine-tune mismatch. 
\addtolength{\tabcolsep}{0pt}
\begin{table*}[t!]
\small
\begin{center}
\begin{tabularx}{\linewidth}{X l l l l l}
\ttop
 \textbf{Model} & ELECTRA & All-Tokens MLM & Replace MLM & ELECTRA 15\% & BERT \tstrut \tsep
 GLUE score & 85.0 & 84.3 & 82.4 & 82.4 & 82.2 \tstrut \tbottom
\end{tabularx} 
\end{center}
\vspace{-2mm}
\caption{Compute-efficiency experiments (see text for details).}
\vspace{-1mm}
\label{tab:sample}
\end{table*}
\addtolength{\tabcolsep}{0pt}

\begin{figure}[tb]
\begin{center}
\begin{minipage}[c]{0.61\textwidth}
\includegraphics[width=\textwidth]{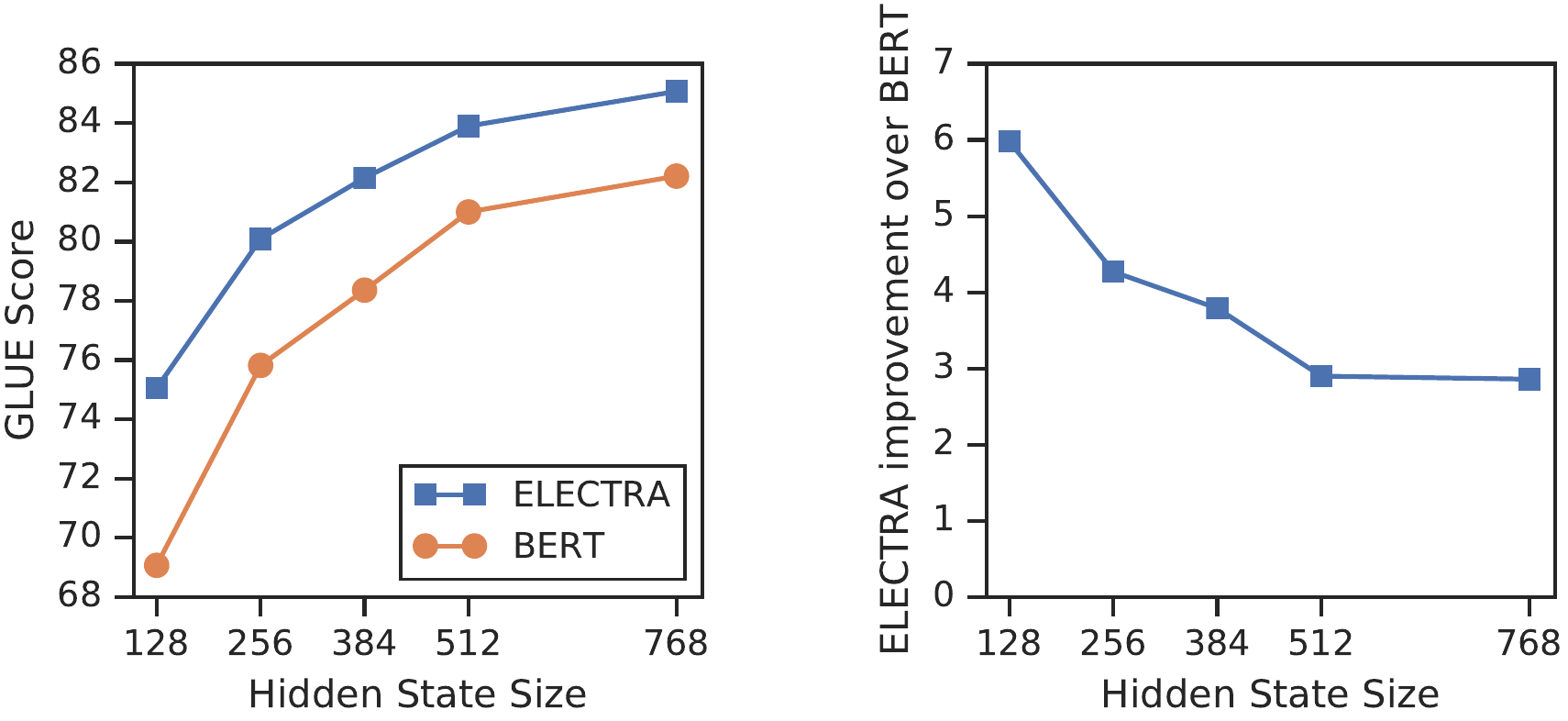}
\end{minipage}
\hspace{6mm}
\begin{minipage}[c]{0.26\textwidth}
\vspace{2mm}
\includegraphics[width=\textwidth]{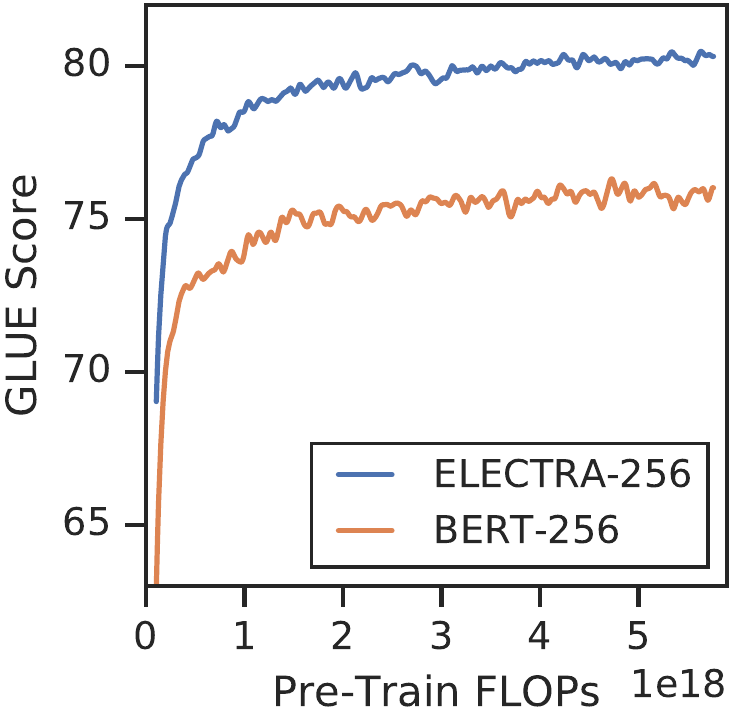}
\end{minipage}
\end{center}
\caption{\underline{Left and Center}: Comparison of BERT and ELECTRA for different model sizes. \underline{\smash{Right}}: A small ELECTRA model converges to higher downstream accuracy than BERT, showing the improvement comes from more than just faster training.}
\label{fig:param}
\end{figure}

The improvement of ELECTRA over All-Tokens MLM suggests that the ELECTRA's gains come from more than just faster training.  
We study this further by comparing BERT to ELECTRA for various model sizes (see Figure~\ref{fig:param}, left).
We find that the gains from ELECTRA grow larger as the models get smaller.
The small models are trained fully to convergence (see Figure~\ref{fig:param}, right), showing that ELECTRA achieves higher downstream accuracy than BERT when fully trained. 
We speculate that ELECTRA is more parameter-efficient than BERT because it does not have to model the full distribution of possible tokens at each position, but we believe more analysis is needed to completely explain ELECTRA's parameter efficiency.

\section{Related Work}

\xhdr{Self-Supervised Pre-training for NLP} 
Self-supervised learning has been used to learn word representations \citep{Collobert2011NaturalLP,pennington2014glove} and more recently {\it contextual} representations of words though objectives such as language modeling \citep{dai2015semi,peters2018deep,howard2018universal}.
BERT \citep{devlin2018bert} pre-trains a large Transformer \citep{Vaswani2017AttentionIA} at the masked-language modeling 
task.
There have been numerous extensions to BERT.
For example, MASS \citep{song2019mass} and UniLM \citep{dong2019unified} extend BERT to generation tasks by adding auto-regressive generative training objectives. 
ERNIE \citep{sun2019ernie} and SpanBERT \citep{joshi2019spanbert} mask out contiguous sequences of token for improved span representations.
This idea may be complementary to ELECTRA; we think it would be interesting to make ELECTRA's generator auto-regressive and add a ``replaced span detection" task.
Instead of masking out input tokens, XLNet \citep{yang2019xlnet} masks attention weights such that the input sequence is auto-regressively generated in a random order. However, this method suffers from the same inefficiencies as BERT because XLNet only generates 15\% of the input tokens in this way.
Like ELECTRA, XLNet may alleviate BERT's pretrain-finetune discrepancy by not requiring $\texttt{[MASK]}$ tokens, although this isn't entirely clear because XLNet uses two ``streams" of attention during pre-training but only one for fine-tuning.
Recently, models such as TinyBERT \citep{jiao2019tinybert} and MobileBERT \citep{sunmobilebert} show that BERT can effectively be distilled down to a smaller model.
In contrast, we focus more on pre-training speed rather than inference speed, so we train ELECTRA-Small from scratch.

\xhdr{Generative Adversarial Networks} GANs \citep{Goodfellow2014GenerativeAN} are effective at generating high-quality synthetic data. 
\citet{radford2015unsupervised} propose using the discriminator of a GAN in downstream tasks, which is similar to our method.
GANs have been applied to text data \citep{Yu2016SeqGANSG,zhang2017adversarial}, although state-of-the-art approaches still lag behind standard maximum-likelihood training \citep{Caccia2018LanguageGF, Tevet2018EvaluatingTG}.
Although we do not use adversarial learning, our generator is particularly reminiscent of MaskGAN \citep{Fedus2018MaskGANBT}, which trains the generator to fill in tokens deleted from the input.

\xhdr{Contrastive Learning} 
Broadly, contrastive learning methods distinguish observed data points from fictitious negative samples.  
They have been applied to many modalities including text \citep{Smith2005ContrastiveET}, images \citep{Chopra2005LearningAS}, and video \citep{Wang2015UnsupervisedLO,Sermanet2017TimeContrastiveNS} data. 
Common approaches learn embedding spaces where related data points are similar \citep{Saunshi2019ATA} or models that rank real data points over negative samples \citep{Collobert2011NaturalLP,Bordes2013TranslatingEF}. 
ELECTRA is particularly related to Noise-Contrastive Estimation (NCE) \citep{Gutmann2010NoisecontrastiveEA}, which also trains a binary classifier to distinguish real and fake data points.

Word2Vec \citep{Mikolov2013EfficientEO}, one of the earliest pre-training methods for NLP, uses contrastive learning.
In fact, ELECTRA can be viewed as a massively scaled-up version of Continuous Bag-of-Words (CBOW) with Negative Sampling.
CBOW also predicts an input token given surrounding context and negative sampling rephrases the learning task as a binary classification task on whether the input token comes from the data or proposal distribution.
However, CBOW uses a bag-of-vectors encoder rather than a transformer and a simple proposal distribution derived from unigram token frequencies instead of a learned generator.

\section{Conclusion}
We have proposed replaced token detection, a new self-supervised task for language representation learning. 
The key idea is training a text encoder to distinguish input tokens from high-quality negative samples produced by an small generator network.
Compared to masked language modeling, our pre-training objective is more compute-efficient and results in better performance on downstream tasks.
It works well even when using relatively small amounts of compute, which we hope will make developing and applying pre-trained text encoders more accessible to researchers and practitioners with less access to computing resources.
We also hope more future work on NLP pre-training will consider efficiency as well as absolute performance, and follow our effort in reporting compute usage and parameter counts along with evaluation metrics.

\section*{Acknowledgements}
We thank Allen Nie, Prajit Ramachandran, audiences at the CIFAR LMB meeting and U. de Montr\'{e}al, and the anonymous reviewers for their thoughtful comments and suggestions. 
We thank Matt Peters for answering our questions about ELMo, Alec Radford for answers about GPT, Naman Goyal and Myle Ott for answers about RoBERTa, Zihang Dai for answers about XLNet, Zhenzhong Lan for answers about ALBERT, and Danqi Chen and Mandar Joshi for answers about SpanBERT. Kevin is supported by a Google PhD Fellowship.

\bibliography{main.bib}
\bibliographystyle{iclr2020_conference}

\appendix

\section{Pre-Training Details}
\label{app:pre}
The following details apply to both our ELECTRA models and BERT baselines.
We mostly use the same hyperparameters as BERT.
We set $\lambda$, the weight for the discriminator objective in the loss to 50.\footnote{As a binary classification task instead of the 30,000-way classification task in MLM, the discriminator's loss was typically much lower than the generator's.}
We use dynamic token masking with the masked positions decided on-the-fly instead of during preprocessing.
Also, we did not use the next sentence prediction objective proposed in the original BERT paper, as recent work has suggested it does not improve scores \citep{yang2019xlnet,liu2019roberta}.
For our ELECTRA-Large model, we used a higher mask percent (25 instead of 15) because we noticed the generator was achieving high accuracy with 15\% masking, resulting in very few replaced tokens.
We searched for the best learning rate for the Base and Small models out of [1e-4, 2e-4, 3e-4, 5e-4] and selected $\lambda$ out of [1, 10, 20, 50, 100] in early experiments. Otherwise we did no hyperparameter tuning beyond the experiments in Section~\ref{sec:extensions}.
The full set of hyperparameters are listed in Table~\ref{tab:hyperpre}. 

\addtolength{\tabcolsep}{0pt}
\begin{table*}[t!]
\begin{center}
\begin{tabularx}{0.9\linewidth}{X l l l}
\ttop
\textbf{Hyperparameter} & \textbf{Small} & \textbf{Base} & \textbf{Large} \tstrut \tsep
Number of layers & 12 & 12 & 24 \tstrut \\
Hidden Size & 256 & 768 & 1024 \\
FFN inner hidden size & 1024 & 3072 & 4096 \\
Attention heads & 4 & 12 & 16 \\
Attention head size & 64 & 64 & 64 \\
Embedding Size & 128 & 768 & 1024 \\
Generator Size (multiplier for hidden-size, & \multirow{2}{*}{1/4} & \multirow{2}{*}{1/3} & \multirow{2}{*}{1/4} \\
FFN-size, and num-attention-heads) & & & \\
Mask percent & 15 & 15 & 25 \\
Learning Rate Decay & Linear & Linear & Linear \\
Warmup steps & 10000 & 10000 & 10000 \\
Learning Rate & 5e-4 & 2e-4 & 2e-4 \\
Adam $\epsilon$ & 1e-6 & 1e-6 & 1e-6 \\
Adam $\beta_1$ & 0.9 & 0.9 & 0.9 \\
Adam $\beta_2$ & 0.999 & 0.999 & 0.999 \\
Attention Dropout & 0.1 & 0.1 & 0.1 \\
Dropout & 0.1 & 0.1 & 0.1 \\
Weight Decay & 0.01 & 0.01 & 0.01 \\
Batch Size & 128 & 256 & 2048 \\
Train Steps (BERT/ELECTRA) & 1.45M/1M & 1M/766K & 464K/400K \tbottom
\end{tabularx} 
\end{center}
\caption{Pre-train hyperparameters. We also train an ELECTRA-Large model for 1.75M steps (other hyperparameters are identical).}
\label{tab:hyperpre}
\end{table*}
\addtolength{\tabcolsep}{0pt}

\section{Fine-Tuning Details}
\label{app:fine}
For Large-sized models, we used the hyperparameters from \citet{Clark2019BAMBM} for the most part.
However, after noticing that RoBERTa \citep{liu2019roberta} uses more training epochs (up to 10 rather than 3) we searched for the best number of train epochs out of [10, 3] for each task. 
For SQuAD, we decreased the number of train epochs to 2 to be consistent with BERT and RoBERTa.
For Base-sized models we searched for a learning rate out of [3e-5, 5e-5, 1e-4, 1.5e-4] and the layer-wise learning-rate decay out of [0.9, 0.8, 0.7], but otherwise used the same hyperparameters as for Large models. 
We found the small models benefit from a larger learning rate and searched for the best one out of [1e-4, 2e-4, 3e-4, 5e-3].
With the exception of number of train epochs, we used the same hyperparameters for all tasks.
In contrast, previous research on GLUE such as BERT, XLNet, and RoBERTa separately searched for the best hyperparameters for each task. 
We expect our results would improve slightly if we performed the same sort of additional hyperparameter search.
The full set of hyperparameters is listed in Table~\ref{tab:hyperfine}.

\addtolength{\tabcolsep}{0pt}
\begin{table*}[t!]
\begin{center}
\begin{tabular}{l l}
\ttop
\textbf{Hyperparameter} & GLUE Value \tstrut \tsep
Learning Rate & 3e-4 for Small, 1e-4 for Base, 5e-5 for Large \tstrut \\
Adam $\epsilon$ & 1e-6  \\
Adam $\beta_1$ & 0.9  \\
Adam $\beta_2$ & 0.999 \\
Layerwise LR decay & 0.8 for Base/Small, 0.9 for Large  \\
Learning rate decay & Linear \\
Warmup fraction & 0.1 \\
Attention Dropout & 0.1  \\
Dropout & 0.1  \\
Weight Decay & 0 \\
Batch Size & 32 \\
Train Epochs & 10 for RTE and STS, 2 for SQuAD, 3 for other tasks 
\tbottom
\end{tabular} 
\end{center}
\caption{Fine-tune hyperparameters}
\label{tab:hyperfine}
\end{table*}
\addtolength{\tabcolsep}{0pt}

Following BERT, we do not show results on the WNLI GLUE task for the dev set results, as it is difficult to beat even the majority classifier using a standard fine-tuning-as-classifier approach.
For the GLUE test set results, we apply the standard tricks used by many of the GLUE leaderboard submissions including RoBERTa \citep{liu2019roberta}, XLNet \citep{yang2019xlnet}, and ALBERT \citep{lan2019albert}. Specifically:
\begin{itemize}
    \item For RTE and STS we use intermediate task training \citep{phang2018sentence}, starting from an ELECTRA checkpoint that has been fine-tuned on MNLI\@. For RTE, we found it helpful to combine this with a lower learning rate of 2e-5. 
    \item For WNLI, we follow the trick described in \citet{liu2019roberta} where we extract candidate antecedents for the pronoun using rules and train a model to score the correct antecedent highly. However, different from \citet{liu2019roberta}, the scoring function is not based on MLM probabilities. Instead, we fine-tune ELECTRA's discriminator so it assigns high scores to the tokens of the correct antecedent when the correct antecedent replaces the pronoun. For example, if the Winograd schema is ``the trophy could not fit in the suitcase because it was too big," we train the discriminator so it gives a high score to ``trophy" in ``the trophy could not fit in the suitcase because the trophy was too big" but a low score to ``suitcase" in ``the trophy could not fit in the suitcase because the suitcase was too big."
    \item For each task we ensemble the best 10 of 30 models fine-tuned with different random seeds but initialized from the same pre-trained checkpoint. 
\end{itemize}
While these tricks do improve scores, they make having clear scientific comparisons more difficult because they require extra work to implement, require lots of compute, and make results less apples-to-apples because different papers implement the tricks differently. We therefore also report results for ELECTRA-1.75M with the only trick being dev-set model selection (best of 10 models), which is the setting BERT used to report results, in Table~\ref{tab:basetest}.

For our SQuAD 2.0 test set submission, we fine-tuned 20 models from the same pre-trained checkpoint and submitted the one with the best dev set score. 

\section{Details about GLUE}
\label{app:glue}
We provide further details about the GLUE benchmark tasks below
\begin{itemize}
    \item \textbf{CoLA:} Corpus of Linguistic Acceptability \citep{Warstadt2018NeuralNA}. The task is to determine whether a given sentence is grammatical or not. The dataset contains 8.5k train examples from books and journal articles on linguistic theory. 
    \item \textbf{SST:} Stanford Sentiment Treebank \citep{Socher2013RecursiveDM}. The tasks is to determine if the sentence is positive or negative in sentiment. The dataset contains 67k train examples from movie reviews.
    \item \textbf{MRPC:} Microsoft Research Paraphrase Corpus \citep{Dolan2005AutomaticallyCA}. The task is to predict whether two sentences are semantically equivalent or not. The dataset contains 3.7k train examples from online news sources.
    \item \textbf{STS:} Semantic Textual Similarity \citep{Cer2017SemEval2017T1}. The tasks is to predict how semantically similar two sentences are on a 1-5 scale. The dataset contains 5.8k train examples drawn from new headlines, video and image captions, and natural language inference data.
    \item \textbf{QQP:} Quora Question Pairs \citep{QQP}. The task is to determine whether a pair of questions are semantically equivalent. The dataset contains 364k train examples from the community question-answering website Quora.
    \item \textbf{MNLI:} Multi-genre Natural Language Inference \citep{Williams2018ABC}. Given a premise sentence and a hypothesis sentence, the task is to predict whether the premise entails the hypothesis, contradicts the hypothesis, or neither. The dataset contains 393k train examples drawn from ten different sources.
    \item \textbf{QNLI:} Question Natural Language Inference; constructed from SQuAD \citep{Rajpurkar2016SQuAD10}. The task is to predict whether a context sentence contains the answer to a question sentence. The dataset contains 108k train examples from Wikipedia.
    \item \textbf{RTE:} Recognizing Textual Entailment \citep{Giampiccolo2007TheTP}. Given a premise sentence and a hypothesis sentence, the task is to predict whether the premise entails the hypothesis or not. The dataset contains 2.5k train examples from a series of annual textual entailment challenges.
\end{itemize}

\section{Further results on GLUE}
\label{app:test}
We report results for ELECTRA-Base and ELECTRA-Small on the GLUE test set in Table~\ref{tab:basetest}. 
Furthermore, we push the limits of base-sized and small-sized models by training them on the XLNet data instead of wikibooks and for much longer (4e6 train steps); these models are called ELECTRA-Base++ and ELECTRA-Small++ in the table. 
For ELECTRA-Small++ we also increased the sequence length to 512; otherwise the hyperparameters are the same as the ones listed in Table~\ref{tab:hyperpre}.
Lastly, the table contains results for ELECTRA-1.75M without the tricks described in Appendix~\ref{app:fine}.
Consistent with dev-set results in the paper, ELECTRA-Base outperforms BERT-Large while ELECTRA-Small outperforms GPT in terms of average score.
Unsurprisingly, the ++ models perform even better.
The small model scores are even close to TinyBERT \citep{jiao2019tinybert} and MobileBERT \citep{sunmobilebert}. These models learn from BERT-Base using sophisticated distillation procedures. 
Our ELECTRA models, on the other hand, are trained from scratch. 
Given the success of distilling BERT, we believe it would be possible to build even stronger small pre-trained models by distilling ELECTRA. 
ELECTRA appears to be particularly effective at CoLA.
In CoLA the goal is to distinguish linguistically acceptable sentences from ungrammatical ones, which fairly closely matches ELECTRA's pre-training task of identifying fake tokens, perhaps explaining ELECTRA's strength at the task.

\addtolength{\tabcolsep}{-4pt}
\begin{table*}[t!]
\small
\begin{center}
\begin{tabularx}{\linewidth}{X l l l l l l l l l l l}
\ttop 
\textbf{Model} & \textbf{Train FLOPs} & \textbf{Params} & \textbf{CoLA} & \textbf{SST} & \textbf{MRPC} & \textbf{STS} & \textbf{QQP} & \textbf{MNLI} & \textbf{QNLI} & \textbf{RTE} & \textbf{Avg.} \tstrut \tsep 
TinyBERT &  6.4e19+ (45x+)  & 14.5M  & 51.1 & 93.1 & 82.6 & 83.7 & 89.1 & 84.6 & 90.4 & 70.0 & 80.6 \tstrut \\
MobileBERT &  6.4e19+ (45x+)  & 25.3M  & 51.1 & 92.6 & 84.5 & 84.8 & 88.3 & 84.3 & 91.6 & 70.4 & 81.0  \\
GPT &  4.0e19 (29x)  & 117M  & 45.4 & 91.3 & 75.7 & 80.0 & 88.5 & 82.1 & 88.1 & 56.0 & 75.9 \\
BERT-Base  &  6.4e19 (45x)  & 110M  & 52.1 & 93.5 & 84.8 & 85.8 & 89.2 & 84.6 & 90.5 & 66.4 & 80.9 \\
BERT-Large &  1.9e20 (135x)  & 335M  & 60.5 & 94.9 & 85.4 & 86.5 & 89.3 & 86.7 & 92.7 & 70.1 & 83.3 \\
SpanBERT      &  7.1e20 (507x)  & 335M  & 64.3 & 94.8 & 87.9 & 89.9 & 89.5 & 87.7 & 94.3 & 79.0 & 85.9 \tsep
ELECTRA-Small &  1.4e18 (1x) & 14M  & 54.6 & 89.1 & 83.7 & 80.3 & 88.0 & 79.7 & 87.7 & 60.8 & 78.0  \tstrut \\
ELECTRA-Small++ &  3.3e19 (18x) & 14M  & 55.6 & 91.1 & 84.9 & 84.6 & 88.0 & 81.6 & 88.3 & 63.6 & 79.7  \\
ELECTRA-Base &  6.4e19 (45x) & 110M    & 59.7 & 93.4 & 86.7 & 87.7 & 89.1 & 85.8 & 92.7 & 73.1 & 83.5 \\
ELECTRA-Base++ &  3.3e20 (182x) & 110M    & 64.6 & 96.0 & 88.1 & 90.2 & 89.5 & 88.5 & 93.1 & 75.2 & 85.7 \\
ELECTRA-1.75M &  3.1e21 (2200x) & 330M    & \textbf{68.1} & \textbf{96.7} & \textbf{89.2} & \textbf{91.7} & \textbf{90.4} & \textbf{90.7} & \textbf{95.5} & \textbf{86.1} & \textbf{88.6}
\tbottom
\end{tabularx} 
\end{center}
\vspace{-1mm}
\caption{Results for models on the GLUE test set. Only models with single-task finetuning (no ensembling, task-specific tricks, etc.) are shown.}
\label{tab:basetest}
\end{table*}
\addtolength{\tabcolsep}{4pt}
\vspace{-1mm}

\section{Counting FLOPs}
\label{app:flops}

We chose to measure compute usage in terms of floating point operations (FLOPs) because it is a measure agnostic to the particular hardware, low-level optimizations, etc. 
However, it is worth noting that in some cases abstracting away hardware details is a drawback because hardware-centered optimizations can be key parts of a model's design, such as the speedup ALBERT \citep{lan2019albert} gets by tying weights and thus reducing communication overhead between TPU workers. 
We used TensorFlow's FLOP-counting capabilities\footnote{See \url{https://www.tensorflow.org/api_docs/python/tf/profiler}} and checked the results with by-hand computation. We made the following assumptions: 
\begin{itemize}
    \item An ``operation" is a mathematical operation, not a machine instruction. For example, an \texttt{exp} is one op like an \texttt{add}, even though in practice the \texttt{exp} might be slower. We believe this assumption does not substantially change compute estimates because matrix-multiplies dominate the compute for most models. Similarly, we count matrix-multiplies as $2*m*n$ FLOPs instead of $m*n$ as one might if considering fused multiply-add operations. 
    \item The backwards pass takes the same number of FLOPs as the forward pass. This assumption is not exactly right (e.g., for softmax cross entropy loss the backward pass is faster), but importantly, the forward/backward pass FLOPs really are the same for matrix-multiplies, which is most of the compute anyway.
    \item We assume ``dense" embedding lookups (i.e., multiplication by a one-hot vector). In practice, sparse embedding lookups are much slower than constant time; on some hardware accelerators dense operations are actually faster than sparse lookups.
\end{itemize}

\section{Adversarial Training}
\label{app:adv}
Here we detail attempts to adversarially train the generator instead of using maximum likelihood.
In particular we train the generator $G$ to maximize the discriminator loss $\lossds$.
As our discriminator isn't precisely the same as the discriminator of a GAN (see the discussion in Section~\ref{sec:method}), this method is really an instance of Adversarial Contrastive Estimation \citep{Bose2018AdversarialCE} rather than Generative Adversarial Training.
It is not possible to adversarially train the generator by back-propagating through the discriminator (e.g., as in a GAN trained on images) due to the discrete sampling from the generator, so we use reinforcement learning instead. 

Our generator is different from most text generation models in that it is non-autogregressive: predictions are made independently. 
In other words, rather than taking a sequence of actions where each action generates a token, the generator takes a single giant action of generating all tokens simultaneously, where the probability for the action factorizes as the product of generator probabilities for each token. 
To deal with this enormous action space, we make the following simplifying assumption: that the discriminator's prediction $D(\bfx, t)$ depends only on the token $x_t$ and the non-replaced tokens $\{x_i: i \not\in \I \}$, i.e., it does not depend on other generated tokens $\{\hat{x}_i: i \in \I \land i \neq t\}$. This isn't too bad of an assumption because a relatively small number of tokens are replaced, and it greatly simplifies credit assignment when using reinforcement learning. 
Notationally, we show this assumption by (in a slight abuse of notation) by writing $D(\hat{x}_t|\bmx)$ for the discriminator predicting whether the generated token $\hat{x}_t$ equals the original token $x_t$ given the masked context $\bmx$. 
A useful consequence of this assumption is that the discriminator score for non-replaced tokens ($D(x_t|\bmx)$ for $t\not\in \I$) is independent of $p_G$ because we are assuming it does not depend on any replaced token. Therefore these tokens can be ignored when training $G$ to maximize $\lossds$. 
During training we seek to find 
\alns{
\argmax_{\theta_G} \lossds = \argmax_{\theta_G} \E_{\bx, \I, \hbx} \Bigg(\sum_{t=1}^n -&\mathbbm{1}(\fx_t = x_t)\log \D(\bfx, t) - \\ &\mathbbm{1}(\fx_t \neq x_t)\log(1 - \D(\bfx, t)) \Bigg)
}
Using the simplifying assumption, we approximate the above by finding the argmax of 
\alns{
&\E_{\bx, \I, \hbx} \Bigg(\sum_{t \in \I}- \mathbbm{1}(\hat{x}_t = x_t)\log \D(\hat{x}|\bmx) - \mathbbm{1}(\hat{x}_t \neq x_t)\log(1 - \D(\hat{x}|\bmx)) \Bigg) \\
= &\E_{\bx, \I} \sum_{t \in \I} \E_{\hat{x}_t \sim p_G} R(\hat{x}_t, \bx) \\
&\text{where } R(\hat{x}_t, \bx) = \begin{cases}
      -\log \D(\hat{x}_t|\bmx) & \text{if}\ \hat{x}_t = x_t \\
      -\log(1 - \D(\hat{x}_t|\bmx)) & \text{otherwise}
    \end{cases}
}
In short, the simplifying assumption allows us to decompose the loss over the individual generated tokens. 
We cannot directly find $\argmax_{\theta_G}$ using gradient ascent because it is impossible to back-propagate through discrete sampling of $\hat{x}$.
Instead, we use policy gradient reinforcement learning \citep{Williams1992SimpleSG}.
In particular, we use the REINFORCE gradient
\alns{
    \nabla_{\theta_G} \lossds \approx \E_{\bx, \I} \sum_{t \in \I}\E_{\hat{x}_t \sim p_G} \nabla_{\theta_g} \log p_G(\hat{x}_t|\bmx) [R(\hat{x}_t, \bx) - b(\bmx, t)]
}

Where $b$ is a learned baseline implemented as $b(\bmx, t) = -\log \text{sigmoid}(w^T h_G(\bmx)_t)$ 
where $h_G(\bmx)$ are the outputs of the generator's Transformer encoder.
The baseline is trained with cross-entropy loss to match the reward for the corresponding position.
We approximate the expectations with a single sample and learn $\theta_G$ with gradient ascent.
Despite receiving no explicit feedback about which generated tokens are correct, we found the adversarial training resulted in a fairly accurate generator (for a 256-hidden-size generator, the adversarially trained one achieves 58\% accuracy at masked language modeling while the same sized MLE generator gets 65\%).
However, using this generator did not improve over the MLE-trained one on downstream tasks (see the right of Figure~\ref{fig:extensions} in the main paper).

\section{Evaluating ELECTRA as a Masked Language Model}
This sections details some initial experiments in evaluating ELECTRA as a masked language model. 
Using slightly different notation from the main paper, given a context $c$ consisting of a text sequence with one token $x$ masked-out, the discriminator loss can be written as
\alns{
    \lossds = -\sum\limits_{x\in\text{vocab}} \Big( &(1 - \pmask) \pdata(x|c)\log D(x, c) + \text{\hspace{19mm}//unmasked token}\\
    &\pmask\pdata(x|c)p_G(x|c)\log D(x, c) + \text{\hspace{17mm}//generator samples correct token}\\
    &\pmask(1 - \pdata(x|c))p_G(x|c)\log(1 - D(x, c))\Big) \text{\hspace{2mm}//generator samples incorrect token} \\
}
Finding the critical points of this loss with respect to $D$ shows that for a fixed generator the optimal discriminator is
\alns{
    D(x, c) = \pdata(x|c)(a + p_G(x|c))/(a\pdata(x|c) + p_G(x|c))
}
which means
\alns{
    \pdata(x|c) = D(x, c)p_G(x|c) / (a(1 - D(x, c)) + p_G(x|c))
}
where $a = (1 - \pmask)/\pmask$ is the number of unmasked tokens for every masked token.
We can use this expression to evaluate ELECTRA as a masked language model by selecting $\text{argmax}_{x\in\text{vocab}} D(x, c)p_G(x|c) / (a(1 - D(x, c)) + p_G(x|c))$ as the model's prediction for a given context.
In practice, selecting over the whole vocabulary is very expensive, so we instead take the argmax over the top 100 predictions from the generator.\footnote{For ELECTRA-Base, this means the upper-bound for accuracy is around 95\%.}
Using this method, we compared ELECTRA-Base and BERT-Base on the Wikipedia+BooksCorpus dataset.
We found that BERT slightly outperformed ELECTRA at masked language modeling (77.9\% vs 75.5\% accuracy).
It is possible that the assumption of an optimal discriminator, which is certainly far from correct, is harming ELECTRA's accuracy under this evaluation scheme.
However, perhaps it is not too surprising that a model like BERT that is trained specifically for generation performs better at generation while a model with a discriminative objective like ELECTRA is better at being fine-tuned on discriminative tasks. 
We think comparisons of BERT's and ELECTRA's MLM predictions might be an interesting way to uncover more about the differences between ELECTRA and BERT encoders in future work.

\section{Negative Results}
\label{app:neg}

We briefly describe a few ideas that did not look promising in our initial experiments:
\begin{itemize}
    \item We initially attempted to make BERT more efficient by strategically masking-out tokens (e.g., masking our rarer tokens more frequently, or training a model to guess which tokens BERT would struggle to predict if they were masked out). This resulted in fairly minor speedups over regular BERT.  
    \item Given that ELECTRA seemed to benefit (up to a certain point) from having a weaker generator (see Section~\ref{sec:extensions}), we explored raising the temperature of the generator's output softmax or disallowing the generator from sampling the correct token. Neither of these improved results.
    \item We tried adding a sentence-level contrastive objective. For this task, we kept 20\% of input sentences unchanged rather than noising them with the generator. We then added a prediction head to the model that predicted if the entire input was corrupted or not. Surprisingly, this slightly decreased scores on downstream tasks.
\end{itemize}

\end{document}